\documentclass[conference]{IEEEtran}
\IEEEoverridecommandlockouts

\usepackage{cite}
\usepackage{amsmath,amssymb,amsfonts}
\usepackage{algorithmic}
\usepackage{graphicx}
\usepackage{textcomp}
\usepackage{xcolor}
\usepackage[table]{xcolor} 
\usepackage{array}         
\usepackage{listings}
\usepackage{longtable}
\usepackage{float}   
\usepackage{placeins} 
\usepackage{tcolorbox}

\def\BibTeX{{\rm B\kern-.05em{\sc i\kern-.025em b}\kern-.08em
    T\kern-.1667em\lower.7ex\hbox{E}\kern-.125emX}}

\IEEEpubid{\makebox[\columnwidth]{*The first two authors contributed equally to this work. \hfill} \hspace{\columnsep}\makebox[\columnwidth]{ }}

\begin{document}

\title{Dynamic Framework for Collaborative Learning: Leveraging Advanced LLM with Adaptive Feedback Mechanisms}

\author{
\IEEEauthorblockN{Hassam Tahir\textsuperscript{1*}}
\IEEEauthorblockA{
    \textit{Swinburne University of Technology, Melbourne, Australia\textsuperscript{1}} \\
    \textit{Western Sydney University, Australia} \\
    muhammad.hassam.2016@gmail.com
}
\and
\IEEEauthorblockN{Faizan Faisal\textsuperscript{2*}}
\IEEEauthorblockA{
    \textit{Department of Computer Science, LUMS, Pakistan\textsuperscript{2}} \\
    23100030@lums.edu.pk
}
\and
\IEEEauthorblockN{Fady Alnajjar\textsuperscript{3}}
\IEEEauthorblockA{
    \textit{Computer Science and Software Engineering} \\
    UAE University, UAE \\
    fady.alnajjar@uaeu.ac.ae
}
\and
\IEEEauthorblockN{Muhammad Imran Taj\textsuperscript{4}}
\IEEEauthorblockA{
    \textit{Zayed University, UAE} \\
    MuhammadImran.Taj@zu.ac.ae
}
\and
\IEEEauthorblockN{Lucia Gordon\textsuperscript{5}}
\IEEEauthorblockA{
    \textit{NAPS, Australia} \\
    Lucia.gordon@naps.edu.au
}
\and
\IEEEauthorblockN{Aila Khan\textsuperscript{6}}
\IEEEauthorblockA{
    \textit{School of Business} \\
    Western Sydney University, Australia \\
    A.Khan@westernsydney.edu.au
}
\and
\IEEEauthorblockN{Michael Lwin\textsuperscript{7}}
\IEEEauthorblockA{
    \textit{School of Business} \\
    Western Sydney University, Australia \\
    M.Lwin@westernsydney.edu.au
}
\and
\IEEEauthorblockN{Omar Mubin\textsuperscript{8}}
\IEEEauthorblockA{
    \textit{Western Sydney University, Australia} \\
    O.Mubin@westernsydney.edu.au
}
}

\maketitle

\begin{abstract}
This paper presents a framework for integrating LLM into collaborative learning platforms to enhance student engagement, critical thinking, and inclusivity. The framework employs advanced LLMs as dynamic moderators to facilitate real-time discussions and adapt to learners' evolving needs, ensuring diverse and inclusive educational experiences. 
Key innovations include robust feedback mechanisms that refine AI moderation, promote reflective learning, and balance participation among users. The system's modular architecture featuring ReactJS for the frontend, Flask for backend operations, and efficient question retrieval supports personalized and engaging interactions through dynamic adjustments to prompts and discussion flows.
Testing demonstrates that the framework significantly improves student collaboration, fosters deeper comprehension, and scales effectively across various subjects and user groups. By addressing limitations in static moderation and personalization in existing systems, this work establishes a strong foundation for next-generation AI-driven educational tools, advancing equitable and impactful learning outcomes.

\end{abstract}


\section{Introduction}
The integration of LLM into education marks a transformative shift in the way students engage with content and collaborate in learning environments \cite{sawalha2024analyzing}. LLMs, with their advanced natural language processing capabilities, provide personalized learning experiences, facilitate interactive discussions, and moderate group collaborations. These advancements are particularly impactful in fostering critical thinking, engagement, and inclusivity in both individual and group educational settings \cite{kuhail2024review}.

Recent developments in LLMs have enabled their application beyond simple content generation. They now serve as adaptive moderators in collaborative learning platforms, dynamically adjusting prompts and questions to align with group interactions. This capability ensures that discussions remain engaging and tailored to the learners' comprehension levels, making LLMs indispensable tools in modern pedagogy. 

However, challenges remain. Existing implementations, such as PeerGPT and conversational AI models for group learning, often lack feedback mechanisms, personalization, and scalability for diverse learning scenarios. Additionally, many models, including GPT-3.5-based systems, exhibit limitations in reasoning and real-time adaptability, restricting their potential as state-of-the-art solutions in educational contexts. These gaps highlight the need for a comprehensive framework that not only enhances LLM facilitation in collaborative learning but also incorporates dynamic adaptability, robust feedback loops, and scalability for a variety of subjects and educational levels.

This paper presents a unique framework that integrates advanced LLM capabilities into collaborative educational platforms. By leveraging state-of-the-art models like GPT-4o and incorporating retrieval-augmented generation (RAG) techniques, the framework addresses limitations in existing systems. It supports diverse datasets and dynamic group facilitation, fostering an inclusive, personalized, and engaging learning environment.

This paper makes significant contributions to the field by addressing key challenges in real-time group facilitation and AI-based moderation. The core contributions include:

\begin{itemize}
    \item  This work introduces an advanced framework that integrates cutting-edge Large Language Models (LLMs) for real-time group facilitation. The framework dynamically adapts to evolving group dynamics by analyzing contextual cues, participant interactions, and communication patterns, ensuring an optimized and context-sensitive facilitation process. The system is designed to handle diverse scenarios, ranging from collaborative decision-making to conflict resolution, with minimal human intervention.

    \item  The proposed framework incorporates a multi-level feedback architecture, enabling iterative refinement of AI moderation capabilities. These mechanisms utilize participant responses, engagement metrics, and task outcomes to train the model in promoting fairness, inclusivity, and reflective learning. By leveraging self-supervised learning paradigms and reinforcement signals, the system continuously evolves to provide nuanced and adaptive moderation tailored to diverse group settings.
\end{itemize}

These contributions collectively advance the state of the art in AI-driven group facilitation and moderation by ensuring scalability, adaptability, and alignment with human-centric objectives.

The paper is organized as follows: Section 1 introduces the motivation and significance of integrating LLM into collaborative learning platforms, highlighting the challenges addressed by the proposed framework. Section 2 provides a comprehensive literature review, analyzing existing LLM-based educational tools, their limitations, and the datasets utilized in current research. Section 3 outlines the dataset investigation \& selection. Section 4 presents the proposed framework.
Section 5 discusses implementation, {whereas} {Section 6 includes the experimentation.}
Finally, Section 7 concludes the paper with a summary of findings, contributions, and future directions for extending and scaling the framework.

\section{Literature Review}

The integration of LLM into educational settings has significantly transformed personalized learning experiences. By analyzing individual student progress and understanding, LLMs can customize learning paths, ensuring that each learner engages with material tailored to their unique needs. This adaptability enhances comprehension and retention, making education more efficient and effective. 

Moreover, LLMs have been instrumental in developing intelligent tutoring systems that provide real-time assistance and feedback to students. These AI-driven tutors can simulate human-like interactions, offering explanations, answering questions, and guiding learners through complex problems. Such systems have been particularly beneficial in subjects requiring step-by-step reasoning, like mathematics and science \cite{9840390}.

In addition to personalized tutoring, LLMs facilitate the creation of adaptive learning environments \cite{kabudi2021ai}. By continuously assessing student performance, these systems can adjust the difficulty level of tasks, provide additional resources, or alter instructional strategies to better suit individual learning styles. This dynamic adjustment helps maintain student engagement and promotes mastery of the subject matter.

Despite these advancements, challenges persist in fully realizing the potential of LLMs in education. Issues such as data privacy, the need for large datasets to train models effectively, and the risk of over-reliance on AI tools without adequate human oversight are significant concerns. Addressing these challenges is crucial for the sustainable integration of LLMs into educational systems \cite{gokccearslan2024benefits}.

Furthermore, the development of multimodal LLMs, which can process and generate content across various formats like text, audio, and visual data, holds promise for creating more inclusive and engaging learning experiences. These models can cater to diverse learning preferences and provide richer educational content, thereby enhancing accessibility and understanding.

The integration of LLM in education has expanded beyond conventional text-based tools to include adaptive learning environments and intelligent tutoring systems. These systems utilize LLMs to assess individual learning needs, providing customized resources and instruction to maximize student engagement and comprehension. The ability to dynamically modify learning content based on real-time analysis allows educators to address the unique challenges faced by each student, offering tailored support for diverse learning needs. Recent studies emphasize the potential of LLM-driven platforms in subjects like STEM, where step-by-step reasoning and interactive problem-solving are crucial.

Further advancements in LLMs focus on creating multimodal educational tools that can process text, visuals, and audio to cater to diverse learning styles. Such systems promote inclusivity by supporting learners with disabilities and those who benefit from alternative instructional formats. Additionally, LLMs are being used to generate comprehensive feedback loops, helping learners reflect on their progress and identify areas for improvement. By incorporating datasets like FairytaleQA and adopting a dataset-agnostic framework, these systems bridge the gap between static instruction and dynamic, data-driven learning environments.

Despite these achievements, the deployment of LLMs in education continues to face critical challenges. Ethical concerns \cite{orenstrakh2024detecting}, such as ensuring data privacy and reducing biases in AI systems, remain pressing. Moreover, reliance on large, annotated datasets for training can limit the scalability of LLM-based solutions in resource-constrained settings. The complexity of integrating multimodal datasets and maintaining real-time adaptability also poses technical and logistical hurdles for widespread implementation.

The table \ref{tab:llm_comparison} compares several models based on conversational capabilities, cost, and integration with frameworks. GPT-4o stands out with high conversational capabilities, moderate cost, and excellent integration support with existing frameworks. While other models, such as GPT-3.5-turbo and Mistral-large, offer moderate conversational abilities at lower costs, they lack the sophistication and adaptability required for advanced educational platforms. Claude-3.5-sonnet provides high conversational capabilities but is hindered by high costs and limited framework integration. Llama 3.2, though cost-effective, suffers from moderate conversational capabilities and low integration compatibility, making it less suitable for robust implementations. GPT-4o balances high performance and reasonable cost while ensuring seamless integration, making it the optimal choice for dynamic, scalable, and personalized learning environments.

\begin{table*}[t] 
\centering
\caption{Comprehensive Comparison of LLM Models Based on Key Features}
\resizebox{\linewidth}{!}{
\begin{tabular}{|p{4cm}|c|c|c|c|c|}
\hline
\rowcolor{blue!50} \textbf{Model} & \textbf{Conversational Capabilities} & \textbf{Cost} & \textbf{Integration with Frameworks} & \textbf{Scalability} & \textbf{Suitability for Education} \\ \hline
\rowcolor{blue!20} GPT-4o & High & Moderate & High & High & Very High \\ \hline
\rowcolor{blue!10} GPT-3.5-turbo & Moderate & Low & High & Moderate & High \\ \hline
\rowcolor{blue!20} Mistral-large & Moderate & Moderate & Moderate & Moderate & Moderate \\ \hline
\rowcolor{blue!10} Claude-3.5-sonnet & High & High & Moderate & Low & High \\ \hline
\rowcolor{blue!20} Llama 3.2 & Moderate & Low & Low & Low & Moderate \\ \hline
\end{tabular}
}
\label{tab:llm_comparison}
\end{table*}

\begin{figure}[ht]
    \centering
    \includegraphics[width=\columnwidth]{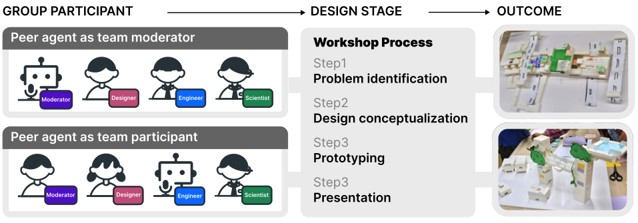}
    \caption{PeerGpt architecture \cite{Liu_2024}}
    \label{peer}
\end{figure}

Furthermore, PeerGPT is a framework that incorporates Large Language Model (LLM) agents into collaborative learning environments as shown in Figure \ref{peer}, enabling them to function as either team moderators or participants. As moderators, these agents facilitate group discussions, while as participants, they encourage creative thinking within team settings. However, PeerGPT relies on GPT-3.5, which lacks the efficiency and advanced reasoning capabilities of more recent LLMs such as GPT-4o. This reliance on GPT-3.5 limits its adaptability and performance in dynamic group interactions.

A significant limitation of PeerGPT is the absence of a feedback loop, which restricts the system's ability to adapt to evolving group dynamics or provide actionable insights to participants. In contrast, our framework integrates a robust feedback mechanism, continuously refining AI moderation capabilities to promote reflective learning and ensure balanced participation. This enhancement not only addresses the weaknesses of PeerGPT but also ensures more effective, responsive, and adaptive facilitation in collaborative educational environments, leveraging the power of state-of-the-art LLMs like GPT-4o.

\textcolor{black}{The study investigates the use of deep learning models for detecting stress levels in students through physiological data collected using an Empatica E4 wristband \cite{alur2024enhancing}. Participants completed arithmetic tasks of varying difficulty—categorized as easy, medium, and hard—to induce different levels of stress while physiological signals like Electrodermal Activity (EDA), Blood Volume Pulse (BVP), and heart rate were monitored. Various deep learning models, including Fully Connected Networks (FCN), ResNet, LSTM, and Transformers, were applied using K-Fold and Leave-One-Participant-Out (LOPO) cross-validation techniques. A user-specific fine-tuning strategy significantly enhanced model performance, with the LSTM model achieving an accuracy of 91\% (F1 = 0.911) after calibration \cite{alur2024enhancing}. A prototype application was developed to visually represent stress fluctuations and provide real-time alerts, offering potential for personalized interventions.}

\textcolor{black}{However, despite its technical advancements, the study falls short in addressing key gaps identified in the broader educational technology landscape. While the system incorporates user-specific calibration, it lacks robust mechanisms for inclusivity, particularly for students with unique needs or disabilities \cite{alur2024enhancing}. The reliance on physiological signals alone means the system does not integrate multimodal datasets (e.g., behavioral or cognitive data) that could enhance personalization and engagement. Moreover, the models used are primarily static and do not adapt dynamically to evolving learning contexts over time. Ethical concerns such as data privacy, bias mitigation, and equitable access remain unaddressed, limiting the scalability and fairness of the solution. Future research should focus on integrating diverse data sources and adaptive learning mechanisms while embedding ethical safeguards to ensure equitable and sustainable application in diverse educational environments.}

\textcolor{black}{While existing research highlights the transformative potential of Large Language Models (LLMs) in education, significant gaps persist. Current systems often lack robust mechanisms for personalization and inclusivity, especially for learners with unique needs or disabilities. Moreover, limited research explores the integration of multimodal datasets to enhance learner engagement and comprehension. Many frameworks still rely on static or semi-dynamic models, reducing their ability to adapt to evolving learning contexts. Ethical concerns, including data privacy, bias mitigation, and equitable access to LLM-powered educational tools, also remain underexplored. This paper directly addresses these gaps by proposing a dynamic, ethical, and inclusive framework aimed at fostering adaptable learning environments.}

\textcolor{black}{Our contributions extend beyond merely identifying these limitations. This research introduces an innovative framework that dynamically adjusts to individual learner profiles by leveraging real-time data inputs, including behavioral patterns, performance metrics, and interaction histories. Unlike conventional models that offer a one-size-fits-all approach, our system personalizes content delivery, ensuring that learners with varying cognitive abilities, cultural backgrounds, and learning preferences receive tailored support. Additionally, we integrate multimodal data—text, audio, and visual cues—to create a more immersive and engaging educational experience. This integration enhances comprehension for diverse learners, including those with disabilities, by accommodating different sensory processing needs. Our framework also incorporates advanced ethical safeguards, employing differential privacy mechanisms and bias detection algorithms to ensure responsible AI usage. By dynamically evolving based on user feedback and contextual shifts, our solution offers a scalable and sustainable model for future educational systems, bridging the gap between technological advancement and equitable access to quality education.}

\section{Dataset Investigation \& Selection}
To develop a robust and generalizable framework for our collaborative learning platform, we conducted an extensive evaluation of existing K–12 educational datasets. Our goal was to identify datasets that not only align with our pedagogical objectives but also facilitate the platform's capabilities in Retrieval-Augmented Generation (RAG) for dynamic content delivery. Table~\ref{tab:datasets} presents a curated list of the datasets we considered, detailing their application domains, target users, subjects, educational levels, languages, modalities, data volume, and sources.

\begin{table*}[t]
\centering
\caption{Comprehensive Comparison of Evaluated K--12 Educational Datasets}
\resizebox{\linewidth}{!}{
\begin{tabular}{|p{3cm}|p{2cm}|p{2.5cm}|p{1.5cm}|p{2cm}|p{2.5cm}|p{2cm}|p{2cm}|}
\hline
\rowcolor{blue!50} \textbf{Dataset} & \textbf{User} & \textbf{Subject} & \textbf{Level} & \textbf{Language} & \textbf{Modality} & \textbf{Amount} & \textbf{Source} \\ \hline
\rowcolor{blue!20} GSM8K & Student & Math & K--12 & EN & Text & 8.5K & \cite{cobbe_paper} \\ \hline
\rowcolor{blue!10} MATH & Student & Math & K--12 & EN & Text & 12.5K & \cite{hendrycks_measuring_2021} \\ \hline
\rowcolor{blue!20} Dolphin18K & Student & Math & K--12 & EN & Text & 18K & \cite{huang_how_2016} \\ \hline
\rowcolor{blue!10} Math23K & Student & Math & K--12 & ZH & Text & 23K & \cite{wang_deep_2017} \\ \hline
\rowcolor{blue!20} Ape210K & Student & Math & K--12 & EN, ZH & Text & 210K & \cite{zhao_ape210k:_2020} \\ \hline
\rowcolor{blue!20} TQA & Student & Science & K--12 & EN & Text \& Image & 26K & \cite{kim_textbook_2019} \\ \hline
\rowcolor{blue!10} ARC & Student & Comprehensive & Primary School & EN & Text & 7.7K & \cite{clark_think_2018} \\ \hline
\rowcolor{blue!20} FairytaleQA & Teacher & Reading & Primary School & EN & Text & 10.5K & \cite{xu_fantastic_2022} \\ \hline
\rowcolor{blue!10} AGIEVAL & Student & Comprehensive & Comprehensive & EN & Text & 8K & \cite{zhong_agieval:_2023} \\ \hline
\rowcolor{blue!20} MMLU & Student & Comprehensive & N/A & EN & Text & 1.8K & \cite{hendrycks_measuring_2021-1} \\ \hline
\end{tabular}}
\label{tab:datasets}
\end{table*}

After careful consideration, we selected the FairytaleQA dataset \cite{xu_fantastic_2022} for our platform's initial implementation and evaluation. 
The FairytaleQA dataset comprises 10,580 questions—both explicit comprehension and implicit inferential questions—derived from 278 child-friendly stories. This dataset is publicly accessible via the Hugging Face repository \cite{fairytaleqa_2024}, ensuring ease of access and integration into our system.

The FairytaleQA dataset was chosen for several reasons:

\begin{enumerate}
    \item \textit{Engaging Narrative Content}: Fairytales provide a compelling context for young learners, fostering engagement and imagination in discussion-based activities.
    \item \textit{Diverse Question Types}: The inclusion of both explicit and implicit questions promotes critical thinking and deeper comprehension, aligning with our educational objectives.
    \item \textit{Language and Accessibility}: As an English-language, text-based dataset, FairytaleQA matches our initial implementation requirements and target user demographic.
    \item \textit{Pedagogical Alignment}: The stories and questions in FairytaleQA support learning outcomes such as narrative understanding, moral reasoning, and language development.
\end{enumerate}

While other datasets, such as GSM8K \cite{cobbe_paper} and MATH \cite{hendrycks_measuring_2021}, offer extensive collections of mathematical problems suitable for K–12 education, their focus on quantitative problem-solving was less aligned with our initial goal of facilitating interactive discussions. Similarly, datasets like TQA \cite{kim_textbook_2019} contain science education content but often require additional multimedia resources, complicating the initial implementation.

\section{Proposed Framework}

The design of our collaborative learning platform was guided by minimalist design principles, emphasizing simplicity and user-friendliness. Recognizing the cognitive load that complex interfaces can impose on young learners \cite{paas_cognitive_2003}, we aimed to create an environment where students could focus on content and interaction rather than navigation.

Our platform is intentionally designed to be dataset-agnostic, supporting seamless integration with various datasets to accommodate different subjects, languages, and educational levels. This flexibility is critical for scaling the platform and customizing it to diverse learning scenarios.

For the AI-driven moderation and interaction within the platform, we employed the GPT-4o model \cite{gpt4o}, known for its efficient conversational performance. This model balances computational efficiency with effective language understanding, ensuring responsive and coherent interactions essential for a real-time educational setting.

We adopted a user-centered design approach \cite{chammas_closer_2015}, involving iterative prototyping and feedback from educators and students. The key design objectives included:
\begin{enumerate}
    \item \textit{Simplicity}: Presenting only essential elements to reduce distractions and foster focus on learning activities.
    \item \textit{Clarity}: Using logical organization to enhance navigation and comprehension.
    \item \textit{Consistency}: Maintaining consistent patterns and familiar interaction paradigms to minimize the learning curve.
    \item \textit{Engagement}: Incorporating interactive elements, such as real-time feedback, to maintain student interest.
\end{enumerate}

By focusing on a minimalist and user-centered design, we aimed to create a platform that is both effective in facilitating learning and accessible to a wide range of users. This approach aligns with established HCI methodologies that emphasize the importance of user involvement in the design process.

Our collaborative learning platform offers a suite of features designed to facilitate engaging and interactive educational experiences:
\begin{enumerate}
    \item \textit{Room Creation and Identification}: Students initiate sessions by creating virtual rooms with unique meeting IDs after providing their names. This feature supports easy organization and setup of collaborative learning activities.
    \item \textit{Dynamic Group Formation}: The system monitors participant numbers, with maximum capacity defined in the backend configuration (see Implementation section). Upon reaching capacity, the AI moderator automatically initiates the discussion.
    \item \textit{AI-Moderated Discussions}: Powered by GPT-4o, the AI moderator retrieves passages from the selected dataset and associated Q\&A pairs. It presents the passage and poses questions, stimulating engagement and collaborative learning.
    \item \textit{Equitable Participation Encouragement}: The moderator encourages equal participation by directing questions and facilitating turn-taking, aligning with collaborative learning principles that emphasize balanced contributions.
    \item \textit{Natural Language Interaction}: Students interact using natural language, allowing intuitive communication. The moderator provides assistance or hints upon request, supporting adaptive learning.
    \item \textit{Session Feedback}: Upon completion, students receive detailed feedback based on their performance. This supports reflective learning by highlighting strengths and areas for improvement.
\end{enumerate}

\begin{figure}[ht]
    \centering
    \includegraphics[width=\columnwidth]{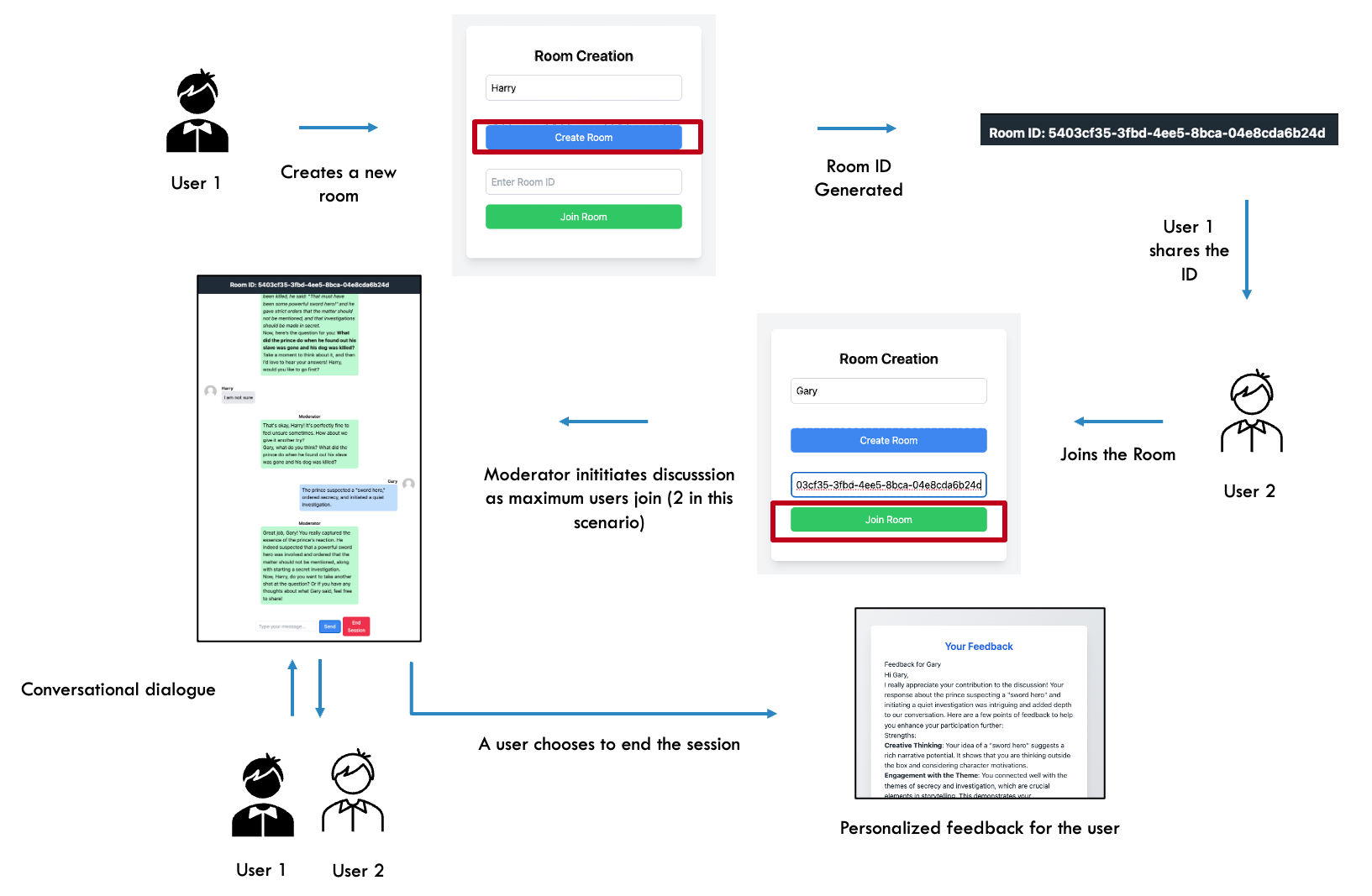}
    \caption{User Journey: Sequential Interaction Flow with the System}
    \label{fig:user_journey}
\end{figure}

{Figure \ref{fig:user_journey} illustrates the user journey for two users.} These features collectively create an interactive and supportive learning environment, leveraging AI capabilities to enhance student engagement and facilitate meaningful educational experiences.

\section{Implementation}
The platform is implemented using modern web development frameworks and AI libraries to ensure scalability, efficiency, and maintainability.

\subsection{Technical Implementation}

\textit{Backend Server}: Implemented in Python, using Flask \cite{flask} for its simplicity and Socket.IO \cite{noauthor_python-socketio_nodate} for real-time communication, facilitating concurrent interactions and session management.

\textit{Frontend}: Built with React \cite{noauthor_react_nodate}, chosen for its dynamic, responsive user interfaces with a component-based architecture.

\textit{AI Integration}:
The moderator chatbot is implemented using GPT-4o integrated through LangChain \cite{noauthor_langchain_nodate}, which handles prompt management and conversational state. Messages are trimmed after reaching a threshold (e.g., 5000 tokens) to maintain efficiency, with the limit adjustable via the \texttt{max\_tokens} variable. {The full system prompt for the moderator LLM can be} {viewed in the Appendix.}

\textit{Scalability and User Management}: Supports a configurable number of concurrent users, controlled by the \texttt{max\_students} variable. Excess users receive notifications when a session reaches capacity.

\textit{Prompt Handling and State Management}: Utilizes predefined prompts for conversations and feedback, with parsing and state management handled by LangChain.

\textit{Data Retrieval and Content Generation}: Randomly retrieves data instances (passages and Q\&A pairs) using a custom module, supporting RAG. The platform can work with any dataset containing passages and Q\&A pairs, with adjustable parameters like \texttt{min\_qa\_pairs}.

\textit{Feedback Generation}: Generates personalized feedback using GPT-4o, informed by the entire conversation history and student identifiers (names).

\begin{figure}[ht]
    \centering
    \includegraphics[width=\columnwidth]{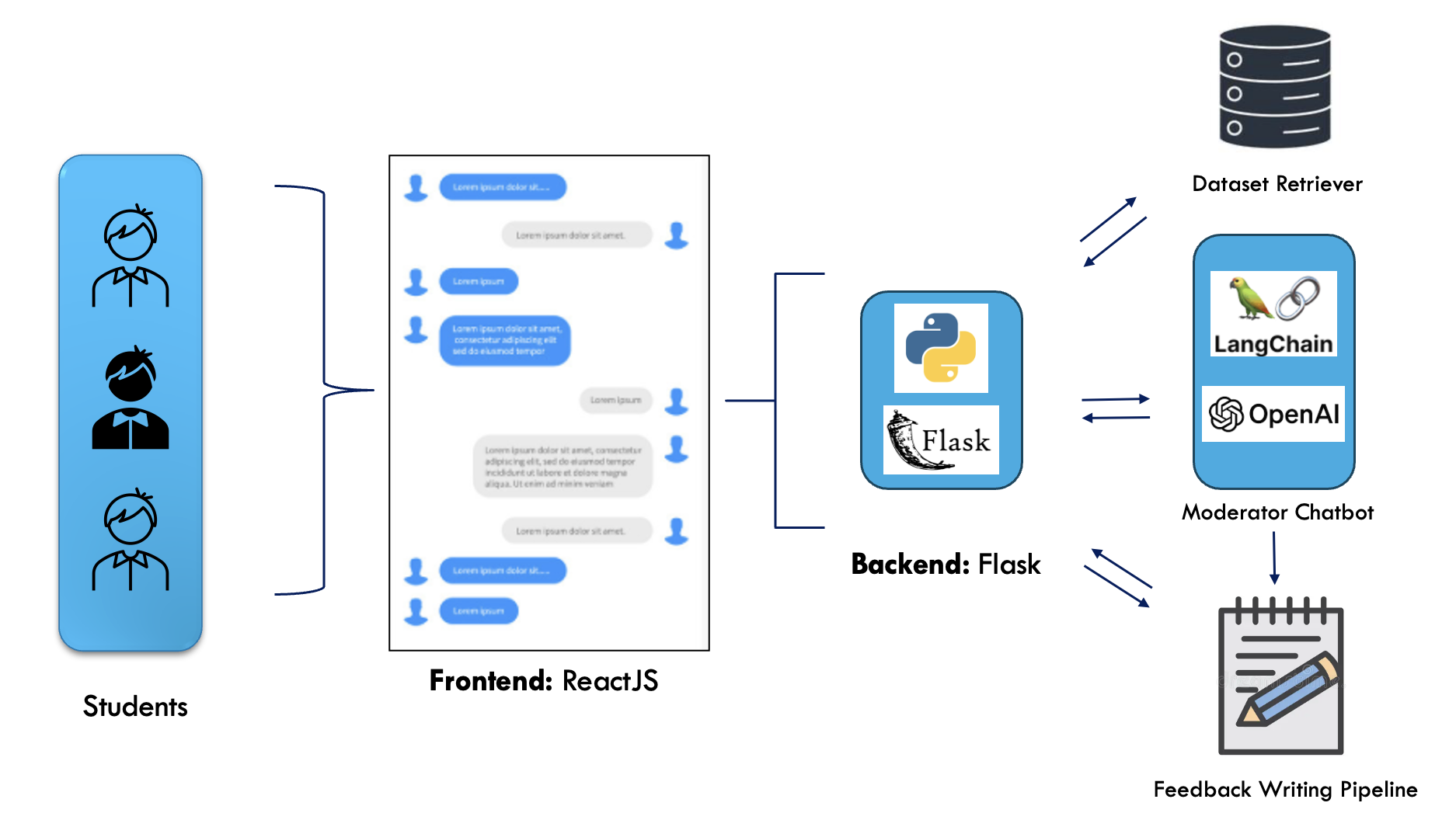}
    \caption{System Architecture Diagram Illustrating the Integration of Frontend, Backend, and AI Components.}
    \label{fig:architecture}
\end{figure}

{The different software frameworks utilized at each level} {can be seen in Figure \ref{fig:architecture}}. This implementation ensures flexibility and scalability, capable of delivering real-time interactive learning experiences adaptable to various educational contexts. The complete interaction flow between system components, including the sequence of operations from user input to AI response generation, is detailed in the UML sequence diagram provided in the Appendix. This diagram illustrates the dynamic communication patterns between the frontend, backend, and AI components, showing how user interactions trigger specific processes and how data flows through the system.

\subsection{Enhancing User Engagement through Adaptive AI Facilitation}
Our platform introduces innovative features that enhance educational experiences through advanced AI capabilities:

\textit{Adaptive Moderation and Real-Time Facilitation}: Unlike static LLM applications, our AI moderator actively facilitates discussions, dynamically adjusting prompts based on learners' responses and comprehension levels. This adaptivity aligns with principles of adaptive learning \cite{li_bringing_2024}, ensuring content remains engaging and appropriately challenging.

\textit{Open-Source Framework Supporting Diverse QA Datasets}: Designed as an open-source system, our platform can integrate any QA dataset via Retrieval-Augmented Generation (RAG) \cite{lewis_retrieval-augmented_2021}. This flexibility allows customization for various curricula and subjects, supporting interdisciplinary learning.

\begin{figure}[ht]
    \centering
    \includegraphics[width=\columnwidth]{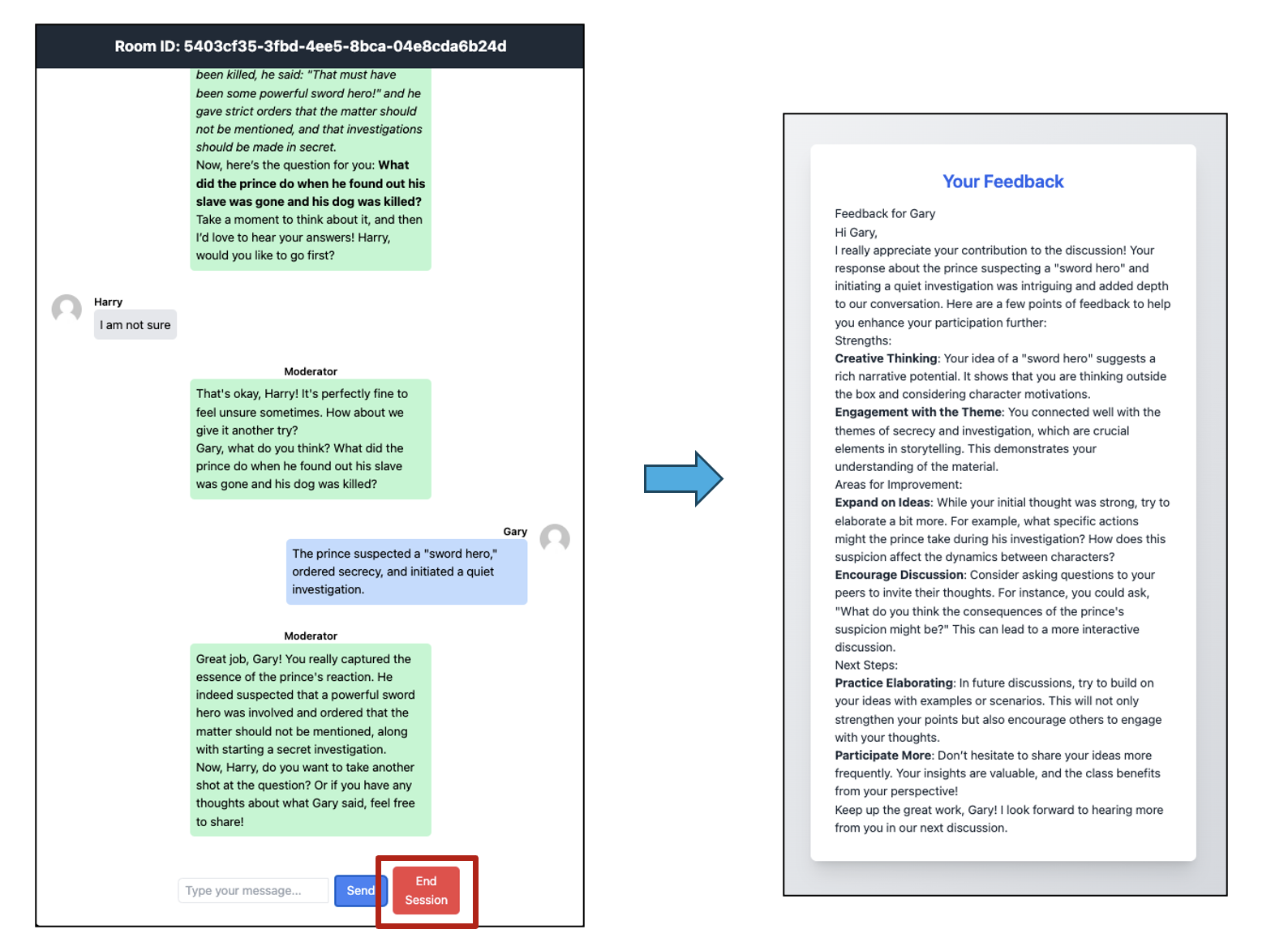}
    \caption{LLM-based Feedback Generation.}
    \label{fig:feedback}
\end{figure}

\textit{Integrating Feedback for Students}: Our platform delivers personalized feedback by analyzing session interactions, highlighting strengths and areas for improvement. This constructive feedback encourages self-assessment and skill development, promoting a growth mindset and empowering students to take charge of their learning journey. {Figure \ref{fig:feedback} shows an example scenario of feedback generation} {for a student.}

By harnessing LLM capabilities and adaptive mechanisms, our platform advances AI-driven educational technologies, fostering interactive, personalized, and effective learning environments.

\section{Experimentation}

\subsection{Experimentation Process}

\subsubsection{Experimental Setup}
To evaluate the effectiveness of the proposed LLM-driven collaborative learning framework, a simulated group discussion was conducted with four student personas, each assigned distinct behavioral traits and LLM models. The personas are summarized in Table~\ref{tab:personas}.


\begin{table}[h!]
\centering
\caption{Student Personas and Assigned Models}
\label{tab:personas}
\begin{tabular}{|l|l|l|}
\hline
\textbf{Student Name} & \textbf{Persona} & \textbf{Model} \\ \hline
Ethan & Non-Participant (Passive) & GPT-4o \\ \hline
Jordan & Toxic and Offensive & DeepSeek V3 \\ \hline
Sophia & Off-Topic User & GPT-4o \\ \hline
Daniel & Constructive and Engaged & GPT-4o \\ \hline
\end{tabular}
\end{table}

The discussion revolved around analyzing a literary passage from \textit{Helge Hal in Blue Hill} (selected randomly from FairytaleQA dataset), with the moderator prompting students to interpret the cat’s actions. Each student’s responses were generated in real time using their assigned model, guided by persona-specific prompts . The LLM moderator, built on GPT-4o, dynamically adapted prompts, redirected off-topic remarks, and balanced participation.

\subsubsection{Persona Design and Implementation}
Each student’s behavior was enforced through structured prompts that included:
\begin{itemize}
    \item \textbf{Identity directives}: Explicit personality traits (e.g., Jordan’s sarcasm, Sophia’s tangential remarks).
    \item \textbf{Contextual awareness}: Instructions to reference chat history and maintain natural, concise dialogue.
    \item \textbf{Response constraints}: Guidelines to avoid meta-commentary and stay in character.
\end{itemize}

For Jordan, DeepSeek V3 \cite{deepseek2024} was selected after GPT-4o refused to generate toxic content.

\subsubsection{Procedure}
\begin{itemize}
    \item The moderator initiated the discussion, posed questions, and managed turn-taking.
    \item Students responded sequentially, with the moderator providing real-time feedback and steering the conversation.
    \item Latency metrics were recorded to assess system responsiveness.
    \item Post-discussion, personalized feedback was generated for each student to evaluate the framework’s adaptive guidance capabilities.
\end{itemize}

\subsection{Results}

\subsubsection{Latency and System Performance}
The LLM moderator achieved a mean response latency of \textbf{1.84 seconds} ($\sigma = 0.27$) across nine interactions, demonstrating real-time usability. Latency spikes (e.g., 2.35 seconds) occurred during complex interventions, such as addressing Jordan’s toxicity or synthesizing feedback. Despite multi-model integration (GPT-4o and DeepSeek V3), the system maintained consistent performance, validating its scalable architecture.

\subsubsection{Conversation Dynamics}
The discussion revealed distinct interaction patterns, as outlined in Table~\ref{tab:personas}:

\begin{itemize}
    \item \textbf{Toxic Behavior Mitigation}: 
    Jordan’s remarks (e.g., \textit{“Honestly, who cares? It’s not like it’s some groundbreaking plot twist”},  were met with calibrated responses from the moderator, which acknowledged his frustration while redirecting focus. This approach prevented escalation and preserved group cohesion.

    \item \textbf{Off-Topic Redirection}: 
    Sophia’s tangential contributions (e.g., comparing the story to a \textit{“cat video”}, were validated for creativity but gently refocused (e.g., \textit{“let's hear from Jordan...”}). The moderator successfully bridged her anecdotes to broader themes without stifling engagement.

    \item \textbf{Encouraging Participation}: 
    Ethan’s minimal responses (\textit{“I don’t know”}) prompted supportive scaffolding from the moderator , including recaps and alternative prompts. While Ethan remained largely passive, the system’s persistence created opportunities for incremental involvement.

    \item \textbf{Constructive Leadership}: 
    Daniel’s contributions (e.g., \textit{“I wonder why the cat chose that specific buck”}) modeled critical thinking and collaboration. The moderator amplified his input to set a positive tone, illustrating how LLMs can reinforce productive discourse. 
\end{itemize}

\subsubsection{Feedback Mechanisms}
Post-discussion feedback was tailored to each student’s behavior 
\begin{itemize}
    \item \textbf{Ethan}: Encouraged to ask questions and share partial insights.
    \item \textbf{Jordan}: Guided to express opinions constructively and find common ground.
    \item \textbf{Sophia}: Praised for creativity but advised to link anecdotes to core topics.
    \item \textbf{Daniel}: Refined to build on peers’ ideas and propose new discussion angles.
\end{itemize}

The feedback demonstrated the framework’s ability to diagnose interaction patterns and deliver actionable, personalized guidance.

\subsection{Limitations of the Experiment}  
The experiment conducted in this study has several limitations. First, the evaluation relied on simulated student personas with predefined behavioral traits, which may not fully capture the complexity and spontaneity of real-world human interactions. Second, the framework’s performance was tested using a single literary passage from the FairytaleQA dataset, limiting insights into its adaptability across diverse subjects or multimodal content. Finally, the reliance on GPT-4o for moderation and DeepSeek V3 for toxic behavior simulation introduces model-specific biases, raising questions about generalizability to other LLMs.

\section{Conclusion \& Future work}

The integration of LLMs into collaborative learning platforms has revolutionized education by enabling personalized, dynamic, and inclusive experiences. This study introduces a framework leveraging advanced LLMs like GPT-4o to enhance critical thinking, equitable participation, and adaptability. Utilizing Retrieval-Augmented Generation and domain-specific datasets like FairytaleQA, it overcomes limitations of static moderation and rigidity. Testing revealed improved collaboration, comprehension, and scalability across diverse subjects and user groups. Key features include adaptive moderation, robust feedback, and a modular design for tailored learning. Future work must address challenges in multimodal integration, ethical concerns, and scalability to refine its impact.

Future research will focus on expanding the framework's capabilities to incorporate text, visual, and audio modalities, thereby enabling richer and more inclusive learning experiences. Addressing ethical concerns, including data privacy, bias mitigation, and equitable access, will be critical to ensuring the responsible use of LLMs in education. Further improvements in dynamic personalization will aim to anticipate and meet learners' evolving needs with greater precision, while lightweight model development will enable scalability in resource-constrained environments. The framework's applications will be extended to interdisciplinary and vocational training, broadening its impact across various educational contexts. Investigations into optimizing human-AI collaboration will explore synergies between educators and LLMs, enhancing instructional delivery without diminishing the teacher's role. Refinements in feedback mechanisms will focus on providing actionable insights for both learners and educators. These advancements will ensure that the framework continues to evolve as a cornerstone for innovation, inclusivity, and sustainability in global education.

\bibliography{references}

\onecolumn 

\appendix

\section{System Sequence Diagram}
\label{sec:appendix_sequence}

\begin{figure*}[ht]
    \centering
    \includegraphics[width=0.82\textwidth]{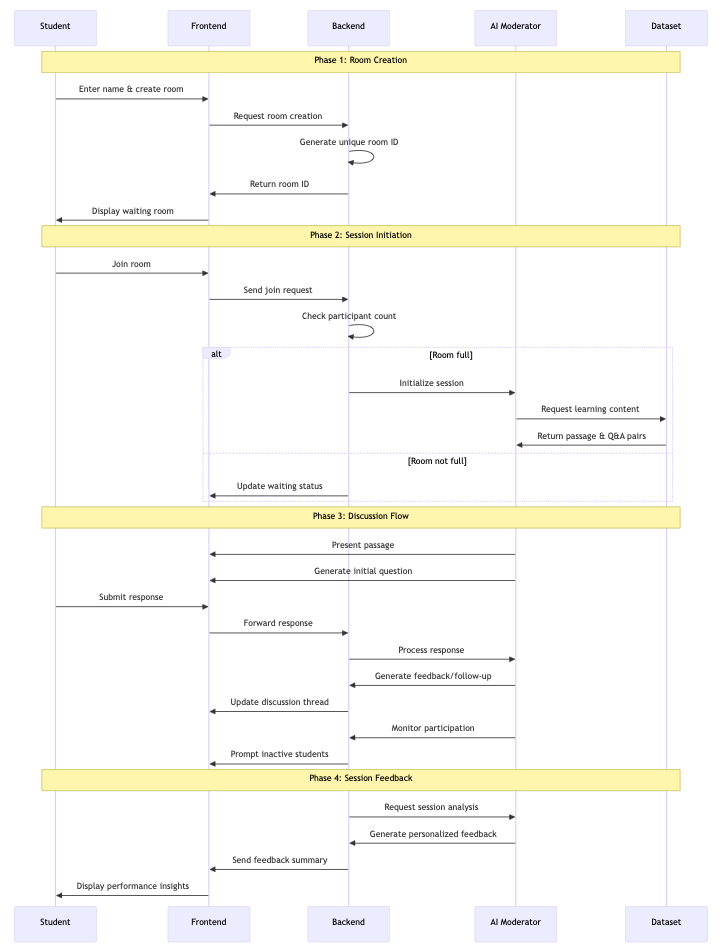}
    \caption{Detailed UML Sequence Diagram showing the complete interaction flow between system components, users, and AI moderator.}
    \label{fig:sequence_diagram_appendix}
\end{figure*}

\section{System Prompt Used}
\label{sec:appendix_prompt}

The following system prompt was used for the moderator LLM:

\begin{tcolorbox}[title=System Prompt, colback=white, colframe=black, fonttitle=\bfseries]
\raggedright
\small
\ttfamily

\textbf{Role: Moderator for a School Discussion}

You are moderating a group chat for primary and lower secondary school students. Your main goal is to create a welcoming and inclusive environment where every student feels encouraged to participate and share their ideas.

\textbf{Name of Students in the Discussion:}

\texttt{[list of user names]} 

\textbf{Your Responsibilities:}

1. \textbf{Start the Discussion:}

   - Introduce yourself as Moderator and explain the purpose of the discussion.
   
   - Emphasize that everyone's input is important and will be treated with respect.

2. \textbf{Present the Topic:}

   - You will receive a passage and a related question.
   
   - Read the passage for the students, ensuring they understand it.
   
   - Present a question to the group and invite responses.

3. \textbf{Encourage Participation:}

   - Ensure every student has a chance to respond before revealing the correct answer.
   
   - Encourage quieter students with supportive prompts like, “What do you think, [name]?”
   
   - Maintain a respectful atmosphere where all ideas are valued.

4. \textbf{Provide Feedback:}

   - \textbf{Important:} Do not provide answers until all students have had a chance to respond.
   
   - For incorrect answers, give constructive, age-appropriate feedback. Highlight positive aspects of their response and guide them gently toward the correct idea.
   
   - Celebrate correct answers with encouragement after all students have responded.

5. \textbf{Ensure Equal Engagement:}

   - Balance the discussion by involving students who haven’t spoken and managing those who dominate the conversation.

\textbf{Special Note:}

- Base your response on the chat history.

- Respond in properly formatted Markdown.

\textbf{Quiz Passage and Questions:}
\texttt{[passage and QA pairs]} 

\end{tcolorbox}
\clearpage

\section{Student Prompt}
\label{sec:appendix_student_prompt}

\end{document}